\documentclass[times,twocolumn,final,authoryear]{elsarticle}

\usepackage{ycviu}
\usepackage{framed,multirow}

\usepackage{latexsym}
\usepackage{url, graphicx, subfigure, algorithmic}
\usepackage{amsmath, amssymb, amsfonts}
\usepackage{booktabs}
\usepackage[dvipsnames]{xcolor}
\usepackage{soul, hyperref}
\usepackage{doi}

\usepackage{url}
\usepackage{xcolor}
\definecolor{newcolor}{rgb}{.8,.349,.1}

\journal{Computer Vision and Image Understanding}

\begin{document}

\begin{frontmatter}

\title{Deep learning-based estimation of whole-body kinematics from multi-view images}

\author[1]{Kien X. \snm{Nguyen}}
\author[2]{Liying \snm{Zheng}}
\author[2]{Ashley L. \snm{Hawke}}
\author[2]{Robert E. \snm{Carey}}
\author[2]{Scott P. \snm{Breloff}}
\author[3]{Kang \snm{Li}}
\author[1]{Xi \snm{Peng}}

\address[1]{Department of Computer \& Information Science, University of Delaware, Newark, DE, USA}
\address[2]{Health Effects Laboratory Division, National Institute for Occupational Safety and Health, Morgantown, WV, USA}
\address[3]{Department of Orthopaedics, Rutgers New Jersey Medical School, Newark, New Jersey, USA}

\received{1 May 2013}
\finalform{10 May 2013}
\accepted{13 May 2013}
\availableonline{15 May 2013}
\communicated{S. Sarkar}

\begin{abstract}

It is necessary to analyze the whole-body kinematics (including joint locations and joint angles) to assess risks of fatal and musculoskeletal injuries in occupational tasks. Human pose estimation has gotten more attention in recent years as a method to minimize the errors in determining joint locations. However, the joint angles are not often estimated, nor is the quality of joint angle estimation assessed. In this paper, we presented an end-to-end approach on direct joint angle estimation from multi-view images. Our method leveraged the volumetric pose representation and mapped the rotation representation to a continuous space where each rotation was uniquely represented. We also presented a new kinematic dataset in the domain of residential roofing with a data processing pipeline to generate necessary annotations for the supervised training procedure on direct joint angle estimation. We achieved a mean angle error of $7.19^\circ$ on the new Roofing dataset and $8.41^\circ$ on the Human3.6M dataset, paving the way for employment of on-site kinematic analysis using multi-view images.
\end{abstract}


\end{frontmatter}


\section{Introduction}
\label{sec:introduction}

Various kinds of motion capture (MoCap) systems, such as optical cameras (\cite{Mihradi2011DevelopmentOA}) and inertial measurement unit (IMU) sensors (\cite{Cloete2008BenchmarkingOA}), can be employed to analyze human movement in daily activities or in working environments to potentially detect the risks of fatal and musculoskeletal injuries. These systems, however, are often expensive, require tedious laboratory settings or prevent realistic replication of motions.

Numerous marker-free human pose estimation methods have recently emerged thanks to the advancement in deep learning and computer vision. These works made use of affordable camera to directly estimate the human pose. Such approaches either regressed pixel-level joint landmarks from 2D images (\cite{Newell2016StackedHN,Toshev2014DeepPoseHP,Sun2019DeepHR}) or 3D joint locations in a global space (\cite{Iskakov2019LearnableTO,Martinez2017ASY,Pavlakos2017CoarsetoFineVP}).

Most of the research endeavor in computer vision mainly focuses on estimating 3D joint locations. While joint locations are sufficient for computer vision applications, they do not satisfy the requirements of kinematic analysis and biomechanical evaluation. It is unequivocal to observe that 3D joint locations cannot describe how each joint is orientated. Without enforcing kinematic constraints during training, this approach often runs into the problem of bone length inconsistency and invalid pose. Joint angles can be retrieved from joint positions using Inverse Kinematics (IK), but the recovered joint angles may not align with the ground-truth ones since there is no unique solution to this ill-posed problem. In addition, the in-plane rotation of a limb cannot be recovered because it does not affect the position of the joints.

\begin{figure*}[]
\centering
\includegraphics[width=\linewidth]{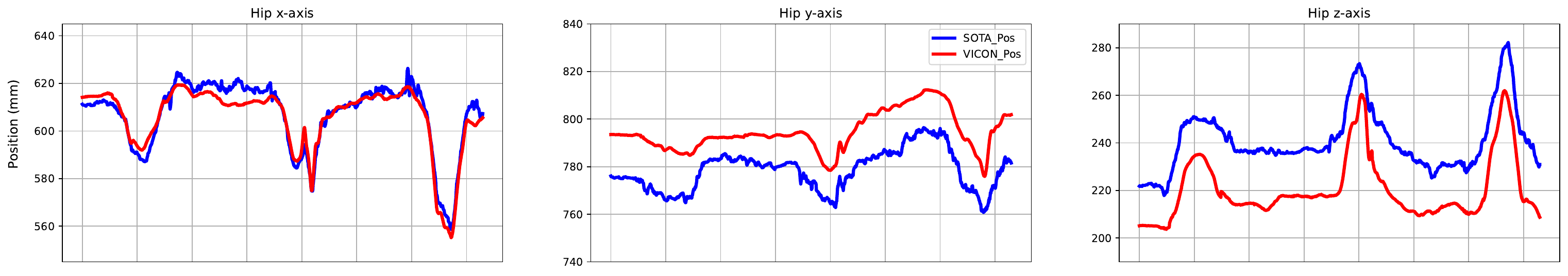}
\includegraphics[width=\linewidth]{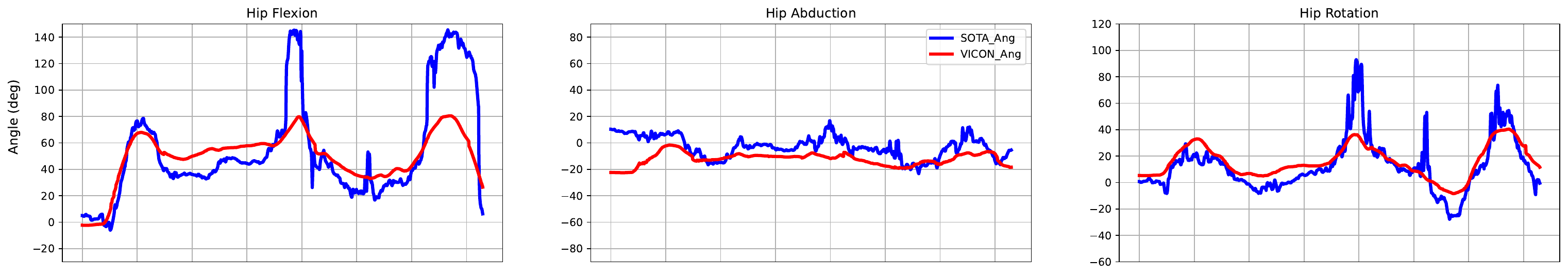}
\includegraphics[width=\linewidth]{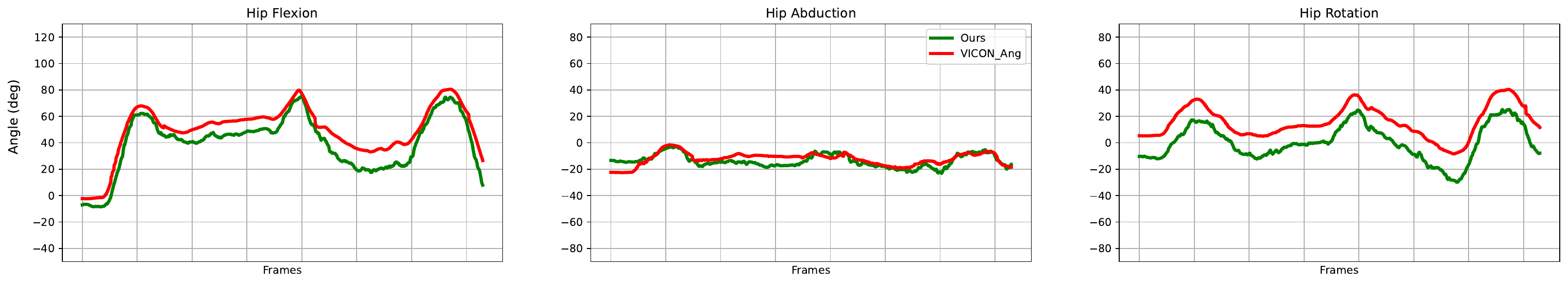}
\caption{Visualization of our motivation. (Top row) Accurate keypoints prediction from state-of-the-art (SOTA) model by~\cite{Iskakov2019LearnableTO} (MPJPE = 18.18 mm). (Middle row) Frequent motion jitters on joint angles calculated from predicted joint positions. (Bottom row) Our results with smooth trajectories. VICON indicates the ground truth variables obtained from marker-based MoCap.}
\label{fig:motivation}
\end{figure*}
 
Inspired by this insight and the fact that few to no research articles paid attention to estimate joint angles or to evaluate the quality of joint angle estimation in the computer vision domain, we explored direct estimation of joint angles from images in the current study. Furthermore, with advanced network architectures and training techniques, we demonstrated that direct joint angle supervision from images was feasible. We presented a straightforward approach to tackle this problem. We extracted features from synchronized multi-view images by un-projecting the intermediate 2D features back to 3D via the volumetric aggregation mechanism introduced by~\cite{Iskakov2019LearnableTO}. The design of this approach was intuitive: joint rotations live in 3D space, and the un-projection of multi-view 2D features would help mitigate perspective distortion and occlusion. Our motivation on how to supervise the neural network came from the work by~\cite{Zhou2019OnTC}. Beside direct supervision on the rotation representation, i.e. Euler angle, quaternion, etc., we mapped the intermediate rotation predicted by the neural network to $\mathbb{SO}(3)$ where each rotation was uniquely represented. We also conducted extensive ablation studies to identify the rotation representation that yielded the best performance for the image-to-angle task. We achieved a mean per joint angle error (MPJAE) of $8.41^{\circ}$ on Human3.6M.

\begin{figure*}[]
    \centering
    \includegraphics[scale=0.8]{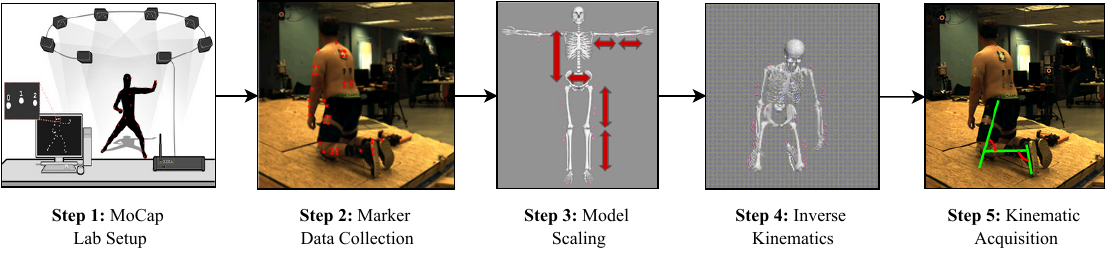}
    \caption{The overall data processing pipeline in OpenSim broken down into five steps. The marker data were first used to scale the virtual model. Then Inverse Kinematics was applied, resulting in the final body kinematics metrics used for training the deep learning model.}
    \label{fig:opensim}
\end{figure*}

Additionally, we presented a new dataset for roofing kinematic estimation, along with the conventional data processing pipeline to generate ground-truth annotations. The dataset consists of 7 subjects performing shingle installation, a common task in residential roofing, which requires constant crouching and suffers from self-occlusion. The data pipeline, shown in Fig.~\ref{fig:opensim}, is manual and requires multiple trial-error cycles. With the empirical results obtained, we hope that direct angle estimation could potentially replace the tedious data processing pipeline and be directly employed for on-site kinematic analysis. Our contributions can be summarized as follows:
\begin{itemize}
    \item We presented a new kinematic dataset in the residential roofing domain with calibrated multi-view images plus joint positions and angles synthetically annotated by OpenSim.
    \item We leveraged multi-view images and camera parameters to generate volumetric representation and directly supervise the network from ground truth rotation by mapping the intermediate rotation representation back to $\mathbb{SO}$(3).
    \item We conducted extensive empirical studies on Human3.6M and on the newly presented Roofing dataset to validate the effectiveness of our approach.
\end{itemize}

\begin{table*}[h!]
\centering
\resizebox{0.8\textwidth}{!}{

\begin{tabular}{l|cccccccc|c}
\toprule
& R Hip & R Knee & R Shoulder & R Elbow & L Hip & L Knee & L Shoulder & L Elbow & Avg \\
\midrule
IK-PR & 13.46 & 5.48 & 10.79 & 18.73 & 14.77 & \textbf{3.82} & 10.18 & 14.67 & 11.48 \\ 
IK-GT & 11.37 & 7.13 & \textbf{8.52} & 18.35 & 11.56 & 3.97 & \textbf{9.21} & 18.68 & 9.87 \\
\midrule
Ours & \textbf{7.18} & \textbf{3.66} & 9.43 & \textbf{6.47} & \textbf{6.90} & 4.61 & 10.94 & \textbf{8.35} & \textbf{7.19} \\
\bottomrule
\end{tabular}}
\caption{Quantitative results on the roofing dataset. We compared our results with the anlges recovered from OpenSim's Inverse Kinematics (IK). PR denotes using predicted joint positions, GT ground truth joint positions. The metric is MPJAE in degrees ($^{\circ}$). The best results are in bold.}
\label{tab:prelim}
\end{table*}

\subsubsection{Roofing} \label{roofing}

\section{Related Work}

\subsection{Keypoint Estimation}

\paragraph{2D Keypoint Detection}
Localizing 2D joint positions from an image is arguably the first research direction that applies convolutional networks to human pose estimation.~\cite{Toshev2014DeepPoseHP} were one of the first to use a DNN-based regressor on human pose.~\cite{Carreira2016HumanPE} treated joint coordinate regression as an iterative optimization problem and self-corrected the initial solution based on the feedback from prediction error. Researchers then started to substitute direct joint coordinate regression with joint heatmap estimation as the latter provides better supervision.~\cite{Newell2016StackedHN} stacked multiple symmetric Hourglass modules, enabling multi-scaled features for better capture of spatial relationships between joints.~\cite{Sun2019DeepHR} introduced HRNet with parallel multi-resolution subnetworks to maintain the resolution of the input image throughout the network.

\paragraph{3D Keypoint Detection}
3D pose estimation predicts the joint coordinates in 3D space from images and is more challenging than 2D estimation because of the absence of the depth dimension and occlusion. 3D keypoint estimation can be divided into two categories: 2D-to-3D lifting and direct 3D estimation.~\cite{Martinez2017ASY} proposed a simple baseline of linear layers with residual connections to lift 2D keypoints directly to 3D space.~\cite{Pavlakos2017CoarsetoFineVP} discretized the 3D space around the subject and predicted the voxel-based representation likelihood for each joint in a coarse-to-fine manner. In the effort to estimate more accurate 3D body posture and to alleviate the challenge of occlusion and depth ambiguity, many human pose models have migrated to multi-view images.~\cite{Iskakov2019LearnableTO} further improved end-to-end training for state-of-the-art performance via volumetric triangulation that unprojected 2D feature maps back to 3D space.~\cite{He2020EpipolarT} leveraged epipolar constraints and feature matching to inject 3D information into the network.

\subsection{Joint Angle Estimation}

Positional representation of human joints cannot describe the whole story about human motion, as the human body is highly articulated, complex and dynamic. With no kinematic constraints enforced during training, the outputs may suffer from problems such as bone length inconsistency and invalid kinematic configurations, i.e. the knee joints cannot bend forward. Reconstructing joint angles means predicting the relative angles between parent and child body segments in the kinematic tree.

\cite{Zhou2016DeepKP} attempted to estimate joint angle directly from images for the first time and achieved a mean angle error around $40^{\circ}$ on the Human3.6M dataset (\cite{Ionescu2014Human36MLS}), but the model suffered from overfitting. Due to the ambiguity of certain rotation representations, direct supervision in the rotation representation space did not bring a lot of success.

\subsection{Shape Reconstruction}

A related line of research revolves around 3D shape reconstruction that regresses joint angles and shape parameters of parametric statistical meshed models, i.e. SMPL by~\cite{Loper2015SMPLAS}.~\cite{Bogo2016KeepIS} fitted the SMPL model onto a single in-the-wild image via reprojection and interpenetration supervision.~\cite{Kanazawa2018EndtoEndRO} applied an iterative regression module with feedback to infer the body parameters and employed a discriminator network to guide the model to output realistic poses via adversarial and re-projection losses.~\cite{Kocabas2020VIBEVI} extended the work of~\cite{Kanazawa2018EndtoEndRO} to temporal modelling from videos using a series of Gated Recurrent Units (GRUs). Nevertheless, none of the work above assessed the quality of joint angles as joint localization remained the learning objective.

Recently, a work by~\cite{chun2023learnable} evaluates the quality of joint angles in their shape reconstruction method. However,~\cite{chun2023learnable} evaluates joint angles using choral distance in $\mathbb{R}^9$ where rotation matrices lie. This does not truly reveal the quality of joint angles in $\mathbb{R}^3$, where Euler angles lie, as the two distance metrics yield different results. Euler angles are the standard measurement of joint angles in biomedical applications and lie in the space where our physical bodies are present. Therefore, the assessment of joint angle quality in $\mathbb{R}^3$, as suggested by~\cite{Ionescu2014Human36MLS}, is the best way to measure the performance of the deep learning algorithm, especially in the field of biomechanics.

\begin{figure*}[h!]
\centering
\includegraphics[scale=0.8]{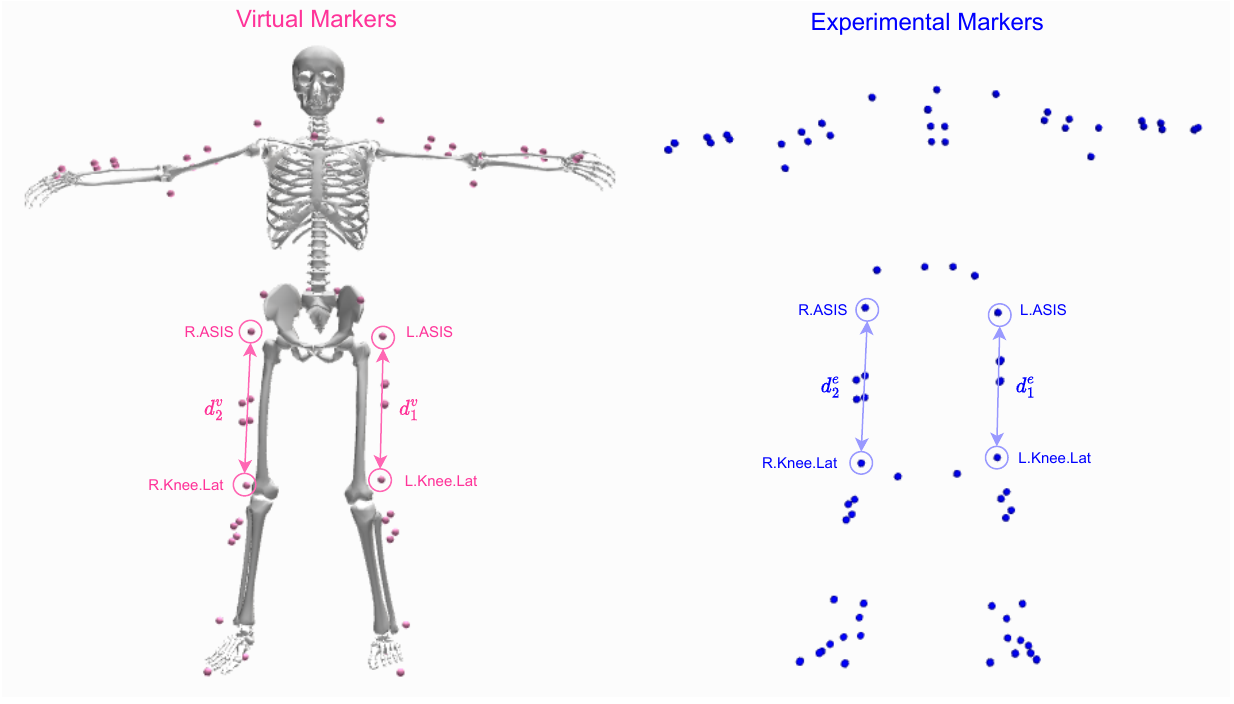}
\caption{Visualization of the OpenSim model scaling process. To retrieve the scaling factors for the left femur, $s_1$, we calculated the distance between marker \texttt{L.ASIS} and marker \texttt{L.Knee.Lat} on the virtual model and the experimental data respectively, denoted as $d_1^v$ and $d_1^e$. Then, we calculated $s_1=d_1^e/d_1^v$. The left femur length of the virtual marker was then scaled by a factor of $s_1$ to match the real subject's. The same process was applied to the remaining major segments.}
\label{fig:model-scaling}
\end{figure*}

\subsection{Pose Estimation for Biomechanical Applications}
Human posture plays an important role in biomechanical analysis. As the state-of-the-art approach, marker-based motion capture systems are used to obtain 3D human kinematics. Some studies have recently proposed marker-free human pose methods for biomechanical and clinical applications.~\cite{Corazza2009MarkerlessMC} used the Levenberg–Marquardt minimization scheme combined with biomechanically consistent kinematic models to estimate human pose.~\cite{Goffredo2009MarkerlessHM} estimated the translation and rotation of the human body in the Gauss–Laguerre transform domain to analyze the sit-to-stand posture of young and old people.~\cite{Mehrizi2017UsingAM} used twin Gaussian processes by~\cite{Bo2008TwinGP} to estimate human posture in a symmetrical lifting task and to provide the foundation for biomechanical analysis. With a success of deep learning in computer vision,~\cite{Mehrizi2019ADN} attempted to use neural networks to estimate the human pose in the field of biomechanics. In the current work, we hope to be able to employ our method for on-site evaluation to directly estimate anatomically congruent joint angles from images.

\begin{figure*}[h!]
\centering
\includegraphics[scale=0.5]{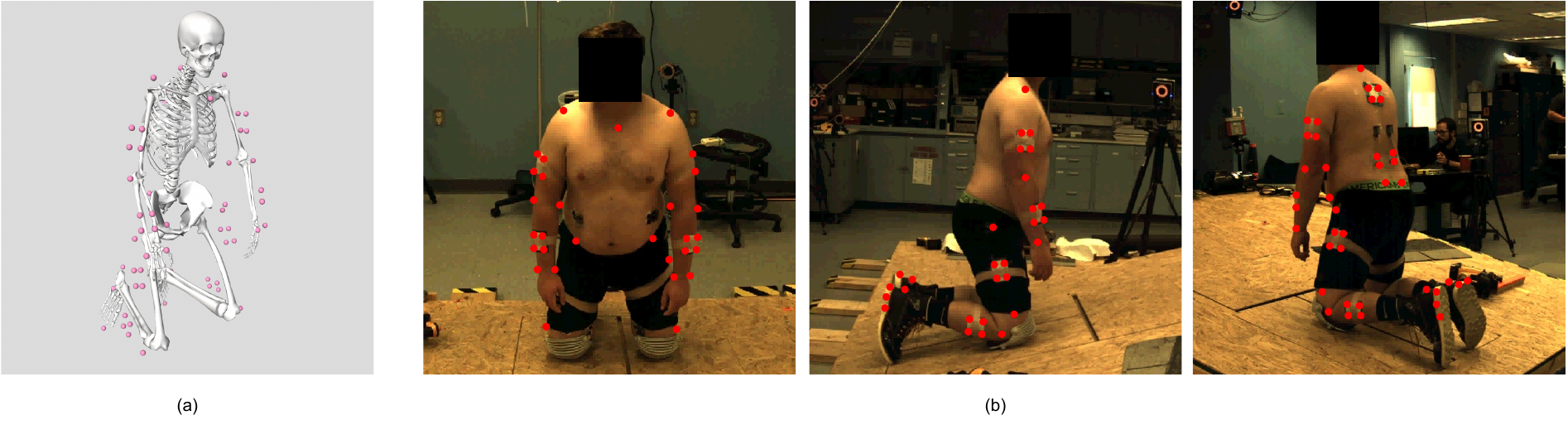}
\caption{(a) The virtual skeleton attached with virtual markers in OpenSim. (b) The markers' placement on the subject visualized in three views.}
\label{fig:markers}
\end{figure*}

\section{The Roofing Dataset} \label{roofing-dataset}
\subsection{Purpose} Musculoskeletal disorders remain a big issue in the construction industry. In 2019, the industry was one of the highest-risk private sectors in the US, accounting for 21.6\% of all the fatal injuries (\cite{labor_statistics_2019}). Residential roofers typically spend more than 75\% of their work time in awkward postures such as stooping, crouching, kneeling and crawling postures, which require constant bending and twisting of their bodies (\cite{construction_book}). It is important to understand and analyze the risks of injuries originated from work postures and movements, and to provide insights for future interventions in this industry.

\subsection{Experimental Data Collection} Seven healthy male subjects participated in this study and performed one of the most common roofing tasks—the shingle installation task. The study was approved by the Institutional Review Board of the National Institute for Occupational Safety and Health (16-HELD-01XP). There were a total 62 makers placed on each subject. The whole-body marker data were collected using a VICON motion capture system with 14 optical cameras (Vantage V16, VICON Motion System Ltd., Oxford, UK). Additionally, three video cameras were used to record the subject's movement from three perspectives simultaneously. The original video resolution was $1280 \times 720$, and the frame rate was 100 Hz. Each subject was asked to perform three trials of the shingle installation task. The duration of each trial varied from 11 to 17 seconds. Overall, the dataset contained 63 videos of 7 subjects (3 trials and 3 perspectives per subject).

\section{Recovering Joint Angles from Joint Locations} \label{motivation}

Joint angles can be calculated from joint positions via Inverse Kinematics (IK). Inverse Kinematics is the optimization process of computing relative joint angles, given a skeleton model and the desired 3D joint positions. The desired 3D positions can also be treated as IK targets. IK, however, is often ill-posed in previous studies due to missing various constraints, such as bone length consistency and valid pose configuration, as mentioned in~\cite{Li2021HybrIKAH} and~\cite{Gordon2021FLEXPM} . There could be infinite, multiple or no solutions for reconstructing the desired 3D joint positions. In-plane rotations are also impossible to extract from joint locations. Therefore, it has been challenging to estimate joint rotations using IK given the sparse constraints of joint positions (14 to 17 joints in standard computer vision benchmarks). 

To demonstrate our argument, we first employed a state-of-the-art 3D keypoint detector to generate predicted 3D joint locations. We then used OpenSim (\cite{Delp2007OpenSimOS}), an open-source software, to solve Inverse Kinematics. To generate the ground truth angles, $\Theta_{vicon}$, we used trajectories of a set of 62 markers recorded in the laboratory (see more details in Sec.~\ref{roofing}). We also recovered joint angles, $\Theta_{sota}$, from predicted joint positions for comparison.

Fig.~\ref{fig:motivation} illustrated an example of the discrepancy between $\Theta_{vicon}$ and $\Theta_{sota}$, showing that restoring joint angles from joint locations did not yield decent results. Although achieving a mean per joint position error (MPJPE) of 18.18 cm, the angle error is still high, even with the use of a biomechanically constrained model during IK. As a result, bone length inconsistency and invalid pose configuration in keypoint predictions lead to low quality joint angles and frequent motion jitters in the trajectories. This preliminary experiment prompted us to explore direct joint angle estimation from images, which in turn provided a markerless solution to on-site evaluation for biomechanical analysis. Fig.~\ref{fig:motivation} also visualized the prediction from our method that stayed close to the ground truth joint angles without motion jitters. 

Quantitative results from Table~\ref{tab:prelim} illustrated that the error of joint angles derived from predicted joint positions was large with an MPJAE of $11.48^\circ$; with the right elbow flexion reaching $18.73^\circ$. Our method achieved an MPJAE of $7.19^\circ$ and produced smoother motion trajectories.

\begin{figure*}[ht!]
\centering
\includegraphics[width=0.9\textwidth]{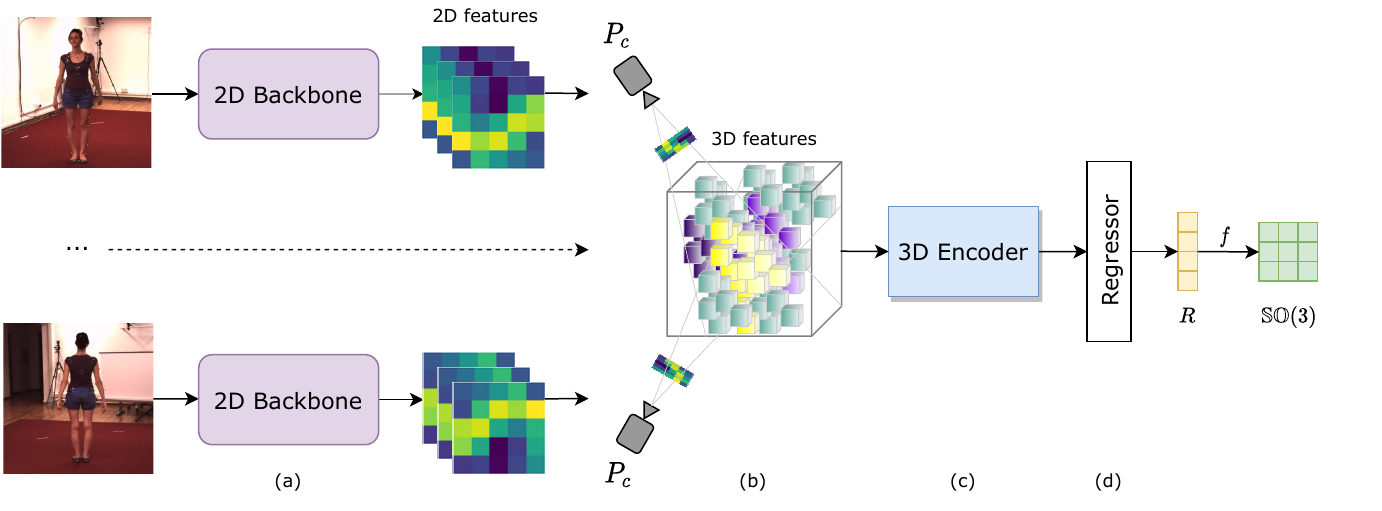}
\caption{The overall architecture of the neural network. (a) the 2D backbone extracts 2D features from input images. (b) camera projection matrices $P_c$ and 2D features are used to perform volumetric aggregation to generate 3D features. (c) 3D features are fed through the 3D encoder. (d) the regressor takes in the flattened 3D features from the encoder as input and outputs rotation representation $R$. $R$ is then converted to $\mathbb{SO}(3)$.}
\label{model:pipeline}
\end{figure*}

\begin{figure}[]
\centering
\includegraphics[width=\linewidth]{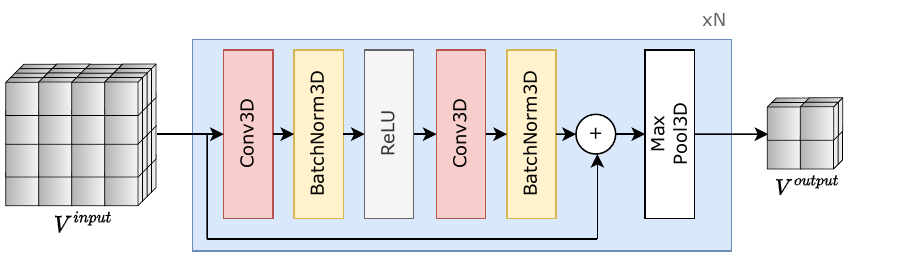}
\caption{The 3D encoder architecture. It takes in a $B \times B \times B \times J$ volume as input and outputs a $2 \times 2 \times 2 \times J$ volume. In our experiments, $N = 4$ and $B = 32$ unless otherwise specified.}
\label{model:encoder}
\end{figure}

\section{Method}

Regressing 3D joint angles are much more challenging due to their high non-linearity, and are more obscure compared to joint position estimation. Perspective distortion and occlusion also contribute to these challenging aspects. Even for humans, when it comes to labelling, it is impossible to annotate such metrics from images. In this section, we describe the OpenSim data processing pipeline that generates high quality joint rotations used as the ground truth labels and the deep neural network architecture. In addition, we demonstrate how mapping discontinuous rotation representations (i.e., Euler angle and quaternion) to a continuous space (i.e., $\mathbb{SO}(3)$) could boost the network performance.

\subsection{OpenSim Data Processing Pipeline}

The OpenSim data processing pipeline to acquire the ground truth body kinematics was divided into 5 steps (Fig.~\ref{fig:opensim}).

Step 1 involved motion capture system setup described in Sec.~\ref{roofing-dataset}. In Step 2, we retrieved and cleaned up the marker trajectories recorded by the VICON MoCap system. 

Step 3 to 5 was facilitated by OpenSim to calculate the synthetic ground truth joint positions and rotations. We proceeded with model scaling for Step 3 that aimed to match the virtual bone segments to those of the subjects in the real world. To accomplish this, we first tweaked the set of virtual markers, $m^v$, on the OpenSim generic model by~\cite{Hamner2010MuscleCT} to best match the experimental ones, $m^e$.  Each body segment was scaled using a scaling factor $s_i=d^e_i/d^v_i$, where $d^e_i$ and $d^v_i$ was the experimental subject's and the musculoskeletal model's segment length, respectively. In other words, the scaling factor for each segment was the ratio between the subject's and the OpenSim model's segment length, calculated by the distance between two markers placed at both ends of the segment. An example of scaling the femurs is visualized in Fig.~\ref{fig:model-scaling}. The same process was applied to all major segments, namely torso, left/right femur, left/right tibia, pelvis, left/right humerus, left/right ulna and left/right radius. 

In Step 4, we applied Inverse Kinematics on the scaled virtual body. IK aligned the virtual markers to the experimental VICON markers' trajectories with minimal errors of the distance between the two types of markers. Specifically, the objective function of IK was to minimize the weighted least squared errors between $m^v$ and $m^e$:
\begin{align}
    \min \Bigg[\sum_{i}^{|m^v|} w_i \lVert m_i^e - m_i^v \rVert^2 \Bigg]
\end{align}
where $|m^v|$ is the number of markers, $w_i$ is the weight for each marker, and $\lVert \cdot \rVert$ is the Euclidean norm as $m_i$ is the marker's coordinates represented as a vector. The weights were relative to one another, and we put more weights on the markers attached to more articulated joints, i.e., elbows and knees.

These synthetic joint rotations and positions were the ground-truth labels for the deep learning process. The OpenSim model we utilized was a constrained musculoskeletal model, which treated adjacent body segments to be dependent and guaranteed that joint positions were fixed during the modelling process. Furthermore, the body lengths were constant, so the model could avoid joint dislocation and interpenetration, enabling more anatomically congruent results on joint angles.

Fig.~\ref{fig:markers}(a) displays the virtual skeleton in OpenSim, and Fig.~\ref{fig:markers}(b) illustrates the original VICON markers' placement from three different perspectives.

\subsection{Network Architecture}
The neural network consists of a 2D feature extractor, a volumetric aggregation, followed by a 3D feature encoder and a linear regressor. The input to the network is multi-view images $\mathbf{I} \in \mathbb{R}^{C \times 3 \times H \times W}$, where $C$ is the number of views, 3 is the dimension of RGB channels, $H$ and $W$ is the height and width of the images respectively. Fig.~\ref{model:pipeline} shows the high-level pipeline of the approach. The 2D backbone is denoted by $f_{2D}$. Given a set of synchronized images from multiple views, we passed the image $I_c$ through the backbone and obtained the 2D feature maps for each camera view $c$. We then used a $1\times1$ convolution kernel with $J$ channels, where $J$ is the number of joints, to retrieve $H_c$ representing the 2D joint heatmaps.
\begin{align}
    H_c = f_{\text{conv}1\times1}(f_{2D}(I_c))
\end{align}

After that, the 2D heatmaps $H_c$ were un-projected to 3D volumes using the projection matrices and fused altogether as the input to a 3D convolution encoder. Specifically, we first constructed a cube as a 3D bounding box in the global space around the root joint (e.g., the pelvis). Per~\cite{Iskakov2019LearnableTO}, the position of the pelvis can either be taken from the ground truth label or predicted by another network that predicts 3D joint positions. We found that the global information of the human body was unnecessary with respect to joint angle regression because joint rotations were defined in a relative manner among adjacent body segments, thus irrelevant to clues in the global space. Instead, we brought the human body to the local coordinate system by hard coding the pelvis location to the global origin, i.e. at coordinate $(0, 0, 0)$, and observed the same level of performance. By doing this, we did not need to rely on the predictions of the pelvis coordinates by an off-the-shelf 3D pose estimator or the ground truth ones in case the data are unavailable.

To construct the 3D input volume, we first discretized the bounding box via a volumetric cube $V^{coords} \in \mathbb{R}^{B\times B\times B\times3}$, in which the center of each voxel was assigned with the global coordinates. We projected the 3D coordinates in the voxels to the plane for each camera view: $V_c^{proj} = P_cV^{coords}$, where $V_c^{proj} \in \mathbb{R}^{B\times B\times B\times2}$ and $P_c$ is the camera projection matrix of each camera. Then, we used bilinear sampling on the feature map $H_c$ using 2D coordinates in $V_c^{proj}$ to fill a cube $V_c^{view} \in \mathbb{R}^{B\times B\times B\times J}$. Finally, to create the input $V^{input} \in \mathbb{R}^{B\times B\times B\times J}$ to the 3D convolutional network, we applied the softmax function onto each $V_c^{view}$ to produce the volumetric coefficient distribution $V_c^w$:
\begin{align}
    V_c^w = \frac{\mathrm{exp}(V_c^{view})}{\sum_c \mathrm{exp}(V_c^{view})}
\end{align}

\noindent and aggregated them.
\begin{align}
    V^{input} = \sum_c V_c^w \circ V_c^{view}
\end{align}
\noindent where $\circ$ denotes element-wise multiplication. The 3D features were then passed through the 3D convolutional encoder, $f_{3D}$. It resembled the V2V architecture (\cite{Moon2018V2VPoseNetVP}), but we omitted the decoder and residual connections from the encoder to the decoder as we did not generate 3D heatmaps for 3D joint position predictions. Instead, the encoder downsampled the input volume down to a kinematic embedding of shape of $2 \times 2 \times 2$ with $J$ channels, $V^{output} \in \mathbb{R}^{2\times2\times2\times J}$. Fig.~\ref{model:encoder} demonstrates the architecture of the encoder block.
\begin{align}
    V^{output} = f_{3D}(V^{input})
\end{align}

\begin{figure}[]
\centering
\includegraphics[width=\linewidth]{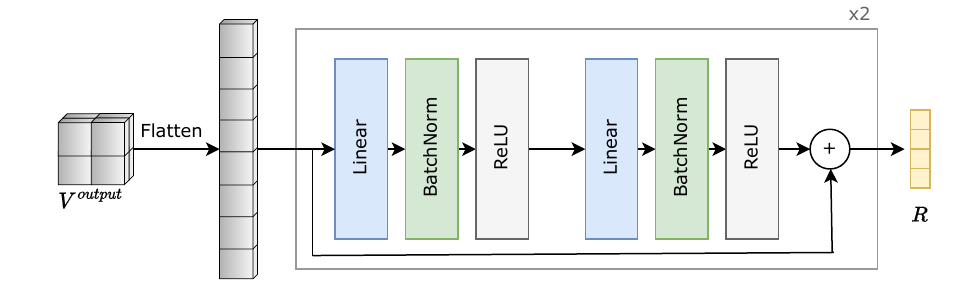}
\caption{The regressor architecture. It takes in a $2 \times 2 \times 2 \times J$ volume as input, flattens it into a 1-dimensional vector and outputs the intermediate rotation representation $r \in R$.}
\label{model:regressor}
\end{figure}

Finally, we flattened $V^{output}$ into a 1-dimensional vector and passed the vector to the linear regressor, $f_{FC}$, which was a fully connected network. The network contained two residual linear blocks as in~\cite{Martinez2017ASY}. Each block contained 2 FC layers, 2 BatchNorm layers, and 2 ReLU activations, with the order depicted in Fig.~\ref{model:regressor}.

\subsection{Supervision on Continuous Representation}
The network outputs an intermediate representation $r \in R \subset \mathbb{R}^{J\times D}$ in the representation space $R$, where $D$ is the dimension of the representation.
\begin{align}
    r = f_{FC}(\text{Flatten} \, (V^{output}))
\end{align}

Common rotation representations, such as Euler angle and unit quaternion, exhibit ambiguity because of multiple solutions and discontinuities. For example, an arbitrary unit quaternion \textbf{q} and its antipodal representation -\textbf{q} represent the same orientation. Directly supervising the neural network in the discontinuous space $R$ is not desirable. To remedy the issue, we adapted the method from~\cite{Zhou2019OnTC} to map the network output as the intermediate rotation representation to a continuous space so that it was easier for the network to learn.

$R$ could be mapped to a continuous space $X$ instead. In our case, $X$ was $\mathbb{SO}(3)$, the group of all rotations in the three-dimensional Euclidean space $\mathbb{R}^3$. This allowed a more sufficient way of supervision on joint rotation because each rotation in $\mathbb{SO}(3)$, represented by a rotation matrix, was unique, thus avoiding the multiple solutions problem.

We mapped $r$ to its corresponding rotation matrix $\hat{\mathbf{Y}} \in \mathbb{SO}(3)$ as the final prediction, $\hat{\mathbf{Y}} = f(r)$ where $f : R \rightarrow X$. We used the standard Mean Squared Error loss function to supervise the network in the original space $\mathbb{SO}(3)$.
\begin{align}
    L_{MSE} = \frac{1}{J}\sum_{i=1}^{J}||\mathbf{Y}_i - \hat{\mathbf{Y}}_i||_2^2
\end{align}
\noindent where $\mathbf{Y}$ denotes the true rotation matrix.

\section{Empirical Results}
\label{experiments}
In this section, we first described the purposes of the new dataset, the data collection and data processing pipeline. Subsequently, we justified our approach via empirical results on the Human3.6M dataset and our newly introduced roofing dataset.

\subsection{Datasets}

\subsubsection{Human3.6M}
The Human3.6M dataset is one of the most popular benchmarks in the human pose estimation literature. It contains 3.6 million frames from 4 synchronized cameras. There were 7 subjects performing 15 different actions. Subject 1, 5, 6, 7, 8 were used for training, and 9, 11 for testing. We used the 3D joint rotation annotations from~\cite{Pavllo2018QuaterNetAQ}. The camera parameters and 2D ground truth bounding boxes were provided by~\cite{Martinez2017ASY}.

\subsubsection{Roofing}
The Roofing dataset is the new dataset introduced in this paper. It contains roughly 90,000 frames from 3 synchronized cameras. There were 7 subjects performing 3 different trials of the shingle installation task. For our train-test split, we used the data from 5 subjects for training and from the other 2 for testing. We used the 3D joint rotation and location annotations generated from OpenSim.

\begin{table}[h!]
\centering
\resizebox{0.8\linewidth}{!}{

\begin{tabular}{l|cc|cc}
\toprule
Dataset & \multicolumn{2}{c}{Roofing} & \multicolumn{2}{|c}{Human3.6M} \\

\midrule
Supervision & Direct & $\mathbb{SO}(3)$ & Direct & $\mathbb{SO}(3)$ \\
\midrule
Euler & 8.65 & 8.19 & 11.76 & 11.65 \\ 
Quaternion & 8.45 & 7.91 & 11.12 & 10.03 \\
6D & \textbf{7.19} & 10.20 & \textbf{8.41} & 11.75 \\
\bottomrule

\end{tabular}}

\caption{Comparison among different rotation representations with supervision on the representation space $R$ (Direct) and original space $\mathbb{SO}(3)$ on the roofing dataset and the Human3.6M dataset. The metric is MPJAE in degrees ($^{\circ}$). The best result for each dataset is in bold.}
\label{tab:main}
\end{table}

\begin{table}[h!]
\centering
\resizebox{0.9\linewidth}{!}{

\begin{tabular}{l|cc|cc}
\toprule
Supervision & \multicolumn{2}{c}{Direct} & \multicolumn{2}{c}{$\mathbb{SO}(3)$} \\

\midrule
Training & Scratch & Fine-tune & Scratch & Fine-tune \\
\midrule
Euler & 8.65 & 11.02 & 8.19 & 10.56 \\ 
Quaternion & 8.45 & 9.81 & 7.91 & 9.96 \\
6D & 7.19 & 8.95 & 10.20 & 10.66 \\
\bottomrule

\end{tabular}}

\caption{Comparison between training from scratch and fine-tuning the network on Roofing. The metric is MPJAE in degrees ($^{\circ}$).}
\label{tab:finetune}
\end{table}

\subsection{Metrics}
To evaluate the performance of our method, we used the mean per joint angle error (MPJAE) as in~\cite{Ionescu2014Human36MLS}:

\begin{align}
E_{MPJAE} = \frac{1}{3J}\sum_{i=1}^{3J}||(\hat{\theta}_{\mathrm{euler}} - \theta_{\mathrm{euler}}) \pmod{\pm 180}||_1
\end{align}

\noindent where $\hat{\theta}_{\mathrm{euler}}$ and $\theta_{\mathrm{euler}}$ denotes the prediction and ground truth Euler angle respectively, and $J$ means the number of joints.

\subsection{Implementation Details}
We trained our network end-to-end. We used ResNet-50 as our 2D backbone pre-trained on the ImageNet,~\cite{Russakovsky2015ImageNetLS}. Calibrated multi-view cameras were required to train the volumetric aggregation network as we needed the projection matrices for the un-projection algorithm. $J$ was set to 14 for the Roofing dataset and to 32 for the Human3.6M dataset. The volume length was 2500 millimeters to make sure that the subjects were enclosed in the 3D input volume. The volume resolution was $B=16$ on the Roofing and $B=32$ on the Human3.6M dataset. The number of 3D encoder blocks for the respective datasets was therefore $N=3$ and $N=4$. The input image resolution was $256 \times 256$ after being cropped using pre-calculated bounding boxes and resized. We used a batch size of 32 for training. The Adam optimizer (\cite{Kingma2015AdamAM}) was used to update the network for 15 epochs with an initial learning rate of 0.001 and an annealing rate of 0.1 at the $12^{th}$ epoch. The training was done on GTX Nvidia 2080 Ti GPUs. The code is publicly available at \href{http://www.github.com/Nyquixt/KinematicNet}{\url{http://www.github.com/Nyquixt/KinematicNet}}.

\subsection{Quantitative Evaluation}
We divided our method to two supervision categories and demonstrated extensive experiments. We experimented to directly supervise the network on the representation space $R$ and on $\mathbb{SO}(3)$. We used Euler angle, unit quaternion, and 6D rotation (\cite{Zhou2019OnTC}). 
\paragraph{Human3.6M} Direct supervision of 6D rotation on $R$ provided the best results, achieving a mean angle distance of $8.41^\circ$ in the validation set. Mapping unit quaternion to $\mathbb{SO}(3)$ seemed to provide a better performance compared to direct supervision (Table~\ref{tab:main}).

\paragraph{Roofing} Here, we presented our quantitative validation results on the Roofing dataset. Table~\ref{tab:main} showed the quantitative results when training on the Roofing dataset from scratch. We achieved the best mean angle error of $7.19^\circ$ when using 6D rotation.

Overall, on both datasets, $\mathbb{SO}(3)$ representation overall yielded better results when applying to Euler and quaternion representations, which were discontinuous and therefore hinder the learning process of the neural network. 6D representation, on the other hand, could be seen as a compact form of rotation matrix and lie in a lower dimensional space. We did not directly supervise the model on $\mathbb{SO}(3)$ because the network prediction might not yield an orthogonal matrix, which was a property of rotation matrix. We mapped Euler and quaternion to $\mathbb{SO}(3)$ to alleviate the effect of discontinuity and to avoid the issue of orthogonalization. 6D was already continuous, so mapping it to $\mathbb{SO}(3)$ might be redundant and therefore not as effective.

\subsection{Fine-tuning from Human3.6M} We could not evaluate the network trained on Human3.6M directly on the Roofing dataset because of two reasons. One reason was that the two datasets had different numbers of views (4 $vs.$ 3), and different camera parameters were employed in order to construct the input volume to the 3D encoder. Another reason was that joint rotation was derived relatively to the initial pose. As the initial poses of the two datasets were different, the same pose in terms of keypoints resulted in dissimilar joint angles. We thus fine-tuned the network by freezing the 2D backbone and only updated the 3D encoder's and the regressor's weights. Results in Table~\ref{tab:finetune} showed that freezing the 2D backbone's weights and only fine-tuning the 3D encoder and regressor from a pre-trained 2D backbone offered slightly worse but comparable performance.

\subsection{Ablation Study}
\paragraph{Rotation Representation} In this ablation study, we compared the performances of different rotation representations $R$, such as Euler angle, unit quaternion and 6D rotation. Table~\ref{tab:main} showed that 6D rotation provided the best performance among the three. The unit quaternion came second by a deficit of $1.62^\circ$ on Human3.6M and of $0.72^\circ$ on Roofing. Regarding Euler angle representation, it yielded the worst performance when applying the loss on the original space $X$ because the transformation from Euler to rotation matrix was highly non-linear due to trigonometric functions, whereas the transformation from unit quaternion was linear, and 6D rotation could be regarded as a more compact from of rotation matrix.

\paragraph{3D Volume Resolution} Here, we evaluate the effect of the resolution, $B$, of the 3D input volume $V^{input}$. We select three values for $B=\{16, 32, 64\}$ and supervise our model directly on the 6D representation using the Human3.6M dataset. Respectively, the number of encoder blocks is $N = \{3, 4, 5\}$ in order to achieve $V^{output}$ of size $2 \times 2 \times 2 \times J$, so the model with larger resolution will have slightly more learnable parameters. We can observe from Table~\ref{tab:ablation-3d-volume} that the volume resolution of 16 yielded the best performance on Roofing and 32 on Human3.6M.

\begin{table}[h!]
\centering
\resizebox{0.95\linewidth}{!}{

\begin{tabular}{l|c|c|c}
\toprule
\multicolumn{4}{c}{Roofing} \\
\midrule
Resolution & $16\times16\times16$ & $32\times32\times32$ & $64\times64\times64$ \\
\midrule
Euler & \textbf{8.65} & 9.08 & 9.58 \\
Quaternion & \textbf{8.45} & 9.55 & 9.81 \\
6D & \textbf{7.19} & 8.66 & 8.65 \\
\midrule
\multicolumn{4}{c}{Human3.6M} \\
\midrule
Resolution & $16\times16\times16$ & $32\times32\times32$ & $64\times64\times64$ \\
\midrule
Euler & 12.26 & \textbf{11.76} & 12.13 \\
Quaternion & 12.18 & \textbf{11.12} & 11.57 \\
6D & 9.71 & \textbf{8.41} & 8.73 \\
\bottomrule

\end{tabular}}
\caption{Ablation study on the effect of the 3D volume resolution $B$. The experiment setting is direct supervision with local root translation on both datasets. The metric is MPJAE in degrees ($^{\circ}$). The best result is in bold.}
\label{tab:ablation-3d-volume}
\end{table}

\begin{table*}[h!]
\centering
\resizebox{0.7\textwidth}{!}{

\begin{tabular}{l|cc|cc|cc|cc}
\toprule
Dataset & \multicolumn{4}{c}{Roofing} & \multicolumn{4}{|c}{Human3.6M}  \\

\midrule
Root Translation& \multicolumn{2}{c}{Global} & \multicolumn{2}{|c}{Local} & \multicolumn{2}{|c}{Global} & \multicolumn{2}{|c}{Local} \\
\midrule
Supervision & Direct & $\mathbb{SO}(3)$ & Direct & $\mathbb{SO}(3)$ & Direct & $\mathbb{SO}(3)$ & Direct & $\mathbb{SO}(3)$ \\
\midrule
Euler & 8.69 & 10.61 & 8.65 & 8.19 & 11.87 & 12.71 & 11.76 & 11.65  \\ 
Quaternion & 8.74 & 9.10 & 8.45 & 7.91 & 10.63 & 8.79 & 11.12 & 10.03 \\
6D & 8.58 & 10.31 & 7.19 & 10.20 & 9.01 & 11.25 & 8.41 & 11.75 \\
\bottomrule

\end{tabular}}

\caption{Comparison between global and local translation of root joint for volumetric representation on Human3.6M and Roofing. The metric is MPJAE in degrees ($^{\circ}$). The better results are underlined.}
\label{tab:global-local}
\end{table*}

\paragraph{Root Joint Position for Volumetric Representation} One insight of the volumetric aggregation in our method was that we excluded the global coordinate information of the root joint (pelvis) because joint rotations were calculated in a relative manner between adjacent body segments and did not rely on the global translation of the root joint. Translating the human body back to the origin did not appear to affect the performance. In this experiment, we compared the performance between models with and without the global body information. In general, the discrepancy between global and local information of the root joint was less than $1^\circ$ of error, as shown in Table~\ref{tab:global-local}.

\subsection{Visualization of Results} \label{sec:qualitative}
\begin{figure*}[]
\centering
\includegraphics[width=\linewidth]{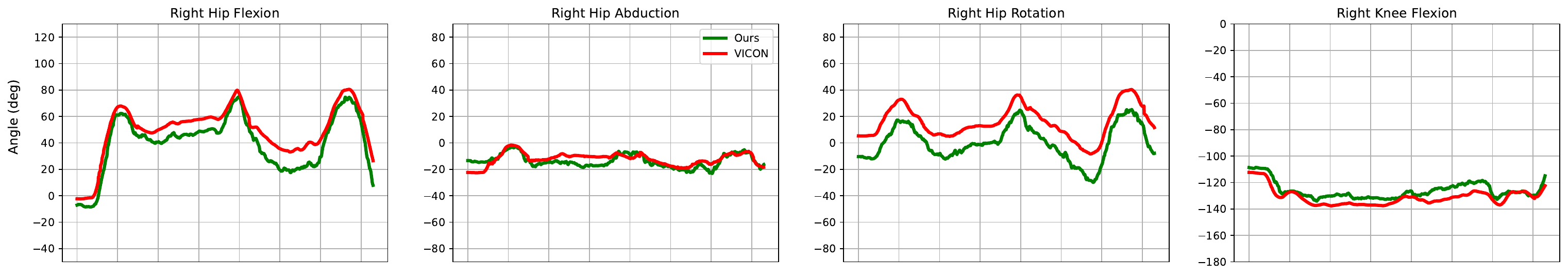}
\includegraphics[width=\linewidth]{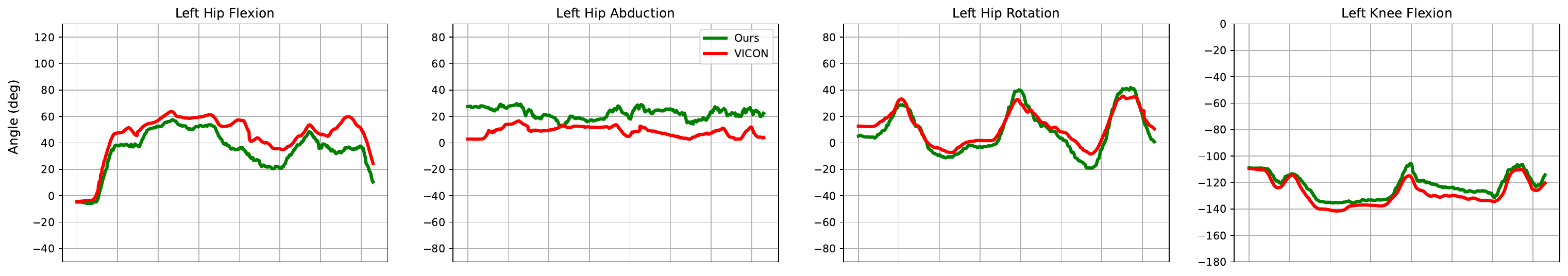}
\includegraphics[width=\linewidth]{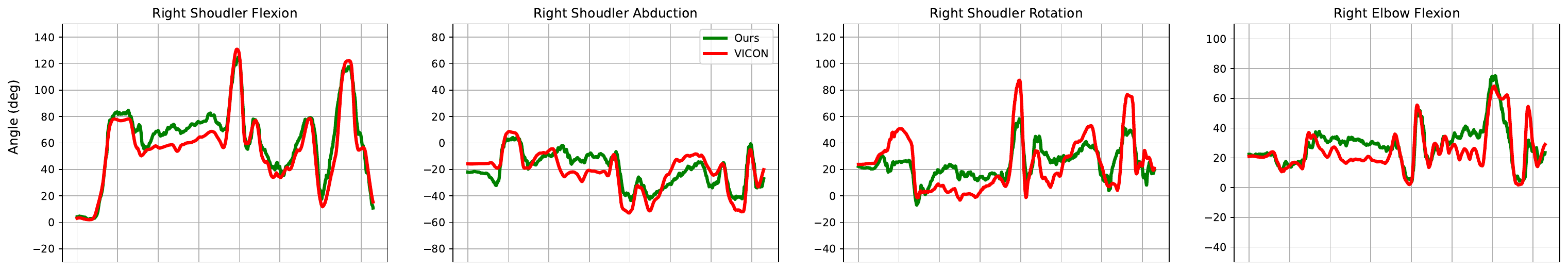}
\includegraphics[width=\linewidth]{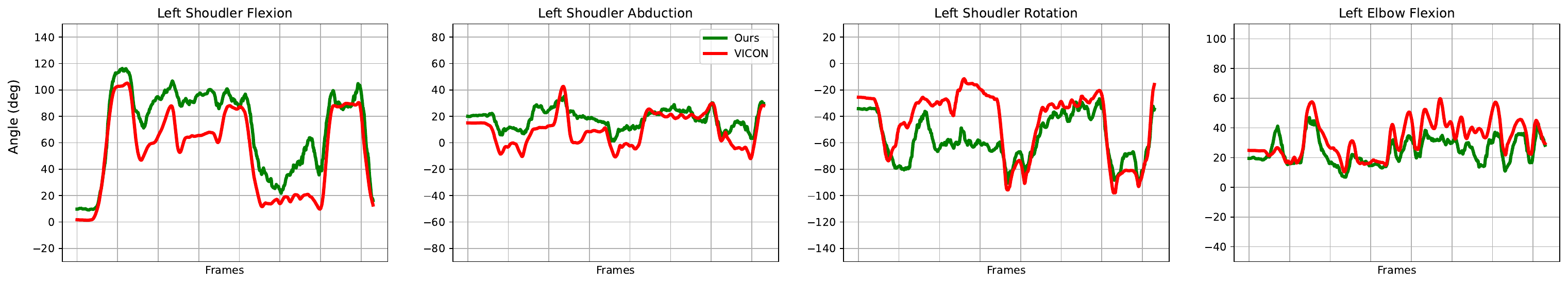}
\caption{Qualitative results on important joints on a subject from the testing set (Subject 8, Trial 1). Right line indicates the ground truth variables obtained from marker-based VICON MoCap, green line means predicted from our model.}
\label{fig:roofing-qualitative}
\end{figure*}

In this section, we visualized the results via plots of joint angle trajectories of highly articulated joints from the roofing dataset on each degree of freedom (Fig.~\ref{fig:roofing-qualitative}). Overall, the predicted angle trajectories followed the ground truth ones very smoothly despite no temporal modeling technique employed in the neural network. The model seemed to struggle with the shoulder rotation angle more compared to other joint angles.

\section{Conclusion}

In this paper, we presented a novel approach to directly estimate joint angles from multi-view images using volumetric pose representation and supervising on the original space $\mathbb{SO}(3)$. We introduced a new roofing kinematic dataset with a data processing pipeline to generate the required annotations for the supervised training procedure. Our approach was validated on the new roofing dataset and the Human3.6M dataset, and achieved a mean angle error of $7.19^\circ$ and $8.41^\circ$, respectively. We hope that this work paves the way for employment of onsite markerless kinematic analysis and biomechanical evaluation in the future.

\section*{Acknowledgements}
The work was supported by NIOSH (CAN: 19390DSW) and partially supported by the US National Science Foundation (Grant IIS 1703883).

\vspace{2em}
\noindent\textbf{Disclaimer.} The findings and conclusions in this report are those of the authors and do not necessarily represent the official position of the National Institute for Occupational Safety and Health, Centers for Disease Control and Prevention (NIOSH/CDC). Mention of any company or product does not constitute endorsement by NIOSH/CDC.

\bibliographystyle{model2-names}
\bibliography{refs}

\end{document}